\documentclass[journal]{IEEEtran}

\usepackage{amsmath,amsfonts}
\usepackage{algorithm}
\usepackage{array}
\usepackage[caption=false,font=normalsize,labelfont=sf,textfont=sf]{subfig} 
\usepackage{textcomp}
\usepackage{stfloats}
\usepackage{url}
\usepackage{verbatim}
\usepackage{graphicx}
\usepackage{cite}
\usepackage{booktabs}
\usepackage{graphicx}
\usepackage{algpseudocode}
\usepackage{multirow}
\usepackage{array}
\usepackage[T1]{fontenc}
\usepackage[colorlinks=false]{hyperref}
\usepackage[capitalise]{cleveref}
\usepackage{titlesec}

\usepackage{subfiles} 

\usepackage{etoolbox}
\makeatletter
\pretocmd\NAT@open{\begingroup\color{\@citecolor}}{}{}
\apptocmd\NAT@close{\endgroup}{}{}
\makeatother

\usepackage[table,xcdraw]{xcolor}
\hypersetup{
    breaklinks=true,
    colorlinks   = true, 
    urlcolor     = black, 
    linkcolor    = blue, 
    citecolor    = blue  
}

\newcommand{\blacklink}[2]{\href{#1}{\textcolor{black}{#2}}}
\newcommand{\blackurl}[1]{\blacklink{#1}{#1}}

\title{Mastering Chess with a Transformer Model}

\newcommand{\orcidlogo}{\includegraphics[height=9pt]{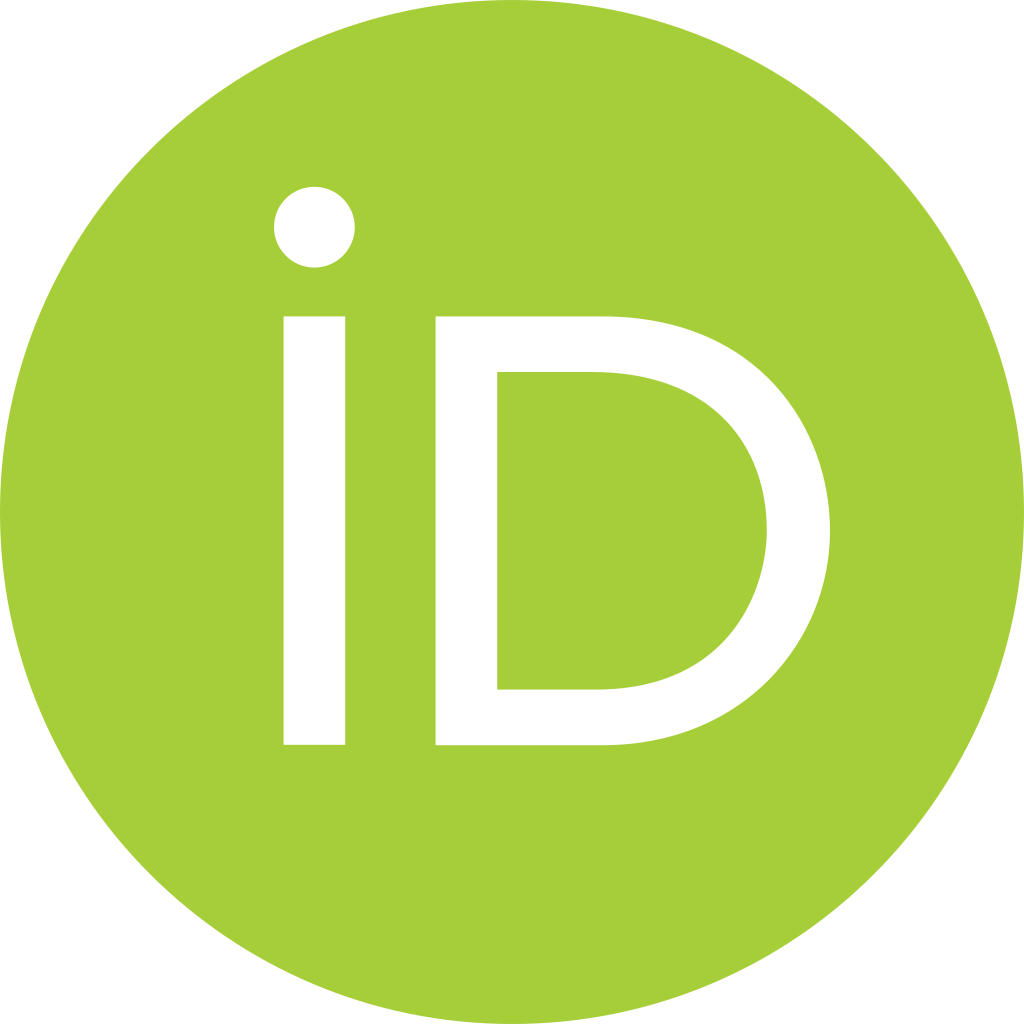}}

\newcommand{\orcidlink}[1]{\href{https://orcid.org/#1}{\orcidlogo}}

\author{Daniel Monroe\orcidlink{0009-0007-1087-1810}, Philip A. Chalmers\orcidlink{0009-0005-1905-7810}

\thanks{\textit{(Daniel Monroe and Philip A. Chalmers are co-first authors.) (Corresponding author: Daniel Monroe)}}
\thanks{Daniel Monroe is with the Department of Mathematics, University of California at San Diego, La Jolla, CA 92093, USA (e-mail: \blacklink{mailto:lc0@danielmonroe.net}{lc0@danielmonroe.net}).}
\thanks{Philip A. Chalmers is with Williams College,  Williamstown, MA 01267, USA. (e-mail:  \blacklink{mailto:pac4@williams.edu}{pac4@williams.edu}).

}

}

\newcommand{\R}{\mathbb{R}}
\newcommand{\newfeature}[1]{\textcolor{blue}{#1}}
\newcommand{\wdl}[3]{({#1}\%W, {#2}\%D, {#3}\%L)}

\begin{document}

\maketitle




\begin{abstract}

Transformer models have demonstrated impressive capabilities when trained at scale, excelling at difficult cognitive tasks requiring complex reasoning and rational decision-making. In this paper, we explore the application of transformers to chess, focusing on the critical role of the position representation within the attention mechanism. We show that transformers endowed with a sufficiently expressive position representation can match existing chess-playing models at a fraction of the computational cost. Our architecture, which we call the Chessformer, significantly outperforms AlphaZero in both playing strength and puzzle solving ability with 8x less computation and matches prior grandmaster-level transformer-based agents in those metrics with 30x less computation. Our models also display an understanding of chess dissimilar and orthogonal to that of top traditional engines, detecting high-level positional features like trapped pieces and fortresses that those engines struggle with. This work demonstrates that domain-specific enhancements can in large part replace the need for model scale, while also highlighting that deep learning can make strides even in areas dominated by search-based methods.

\end{abstract}


\section{Introduction}

\IEEEpubidadjcol

\IEEEPARstart{C}{hess} has long been regarded as a proving ground for artificial intelligence (AI). The importance of strategic planning and rational decision-making in strong chess play make the game an ideal domain in which to test systems aiming to master these human-like cognitive capabilities.

Traditional chess engines employ a specialized tree-search algorithm paired with a handcrafted evaluation function. Most modern chess engines follow this scheme but use an efficiently updated neural network (NNUE) as their evaluation function.

The AlphaZero engine \cite{AlphaZero} introduced a different recipe based on Monte Carlo Tree Search (MCTS) and deep neural networks. This approach was recreated by the open-source Leela Chess Zero (Lc0), which builds on the same principles with several subsequent enhancements. These engines use large networks that predict not only a position evaluation but also a policy vector, a distribution over subsequent moves which is used to guide the search process.

AlphaZero used a convolution-based residual network, which was the state of the art at the time. It was based on AlphaGo, which had prior achieved superhuman performance in Go \cite{AlphaGoZero}. However, as the authors of the AlphaZero paper note, convolution-based models may be poorly suited for chess. Long-range interactions feature much more prominently in chess than in Go, and convolutions are poorly equipped to deal with these because of their small receptive fields. Transformer models, on the other hand, are based on a global self-attention operation and resolve the issue of small receptive
fields. They can also exhibit impressive abilities when trained at scale, powering large language models such as OpenAI's GPT-4 \cite{openai2023gpt4}.

We show that the effectiveness of transformers in chess depends heavily on the choice of position representation in the attention mechanism. Among three representations of varying expressivity, we select for our final architecture the scheme of Shaw et al. \cite{shaw2018selfattention}, training models endowed with this technique and other enhancements at scale. We call the resulting architecture the Chessformer.

Our main contributions are:

\begin{itemize}

\item We present a simple yet performant architecture for transformers in chess and apply this architecture at scale.

\item We ablate the position representation in the attention mechanism of our models to demonstrate the criticality of a sufficiently expressive representation.

\item We provide a detailed comparison of our models against prior work to demonstrate the superiority of our methods.

\item We analyze the attention maps and playstyle of our largest model, CF-240M.

\item We open-source our training code.\footnote{Available on Github at \blackurl{https://github.com/Ergodice/lczero-training}.}

\end{itemize}


This paper is organized as follows. \cref{sec:background} goes over the vanilla self-attention formulation and compares three position representations of varying expressivity. \cref{sec:setup} describes our training setup. \cref{sec:results} compares our models to prior work and reports results for ablating the position representation. \cref{sec:attmaps} analyzes the attention maps of our largest model, CF-240M, while \cref{sec:playstyle} describes its playstyle. \cref{sec:related} reviews related work, and  \cref{sec:conclusion} provides concluding remarks.

\IEEEpubidadjcol

\begin{figure*}
        \begin{center}
        \includegraphics[width=\textwidth]{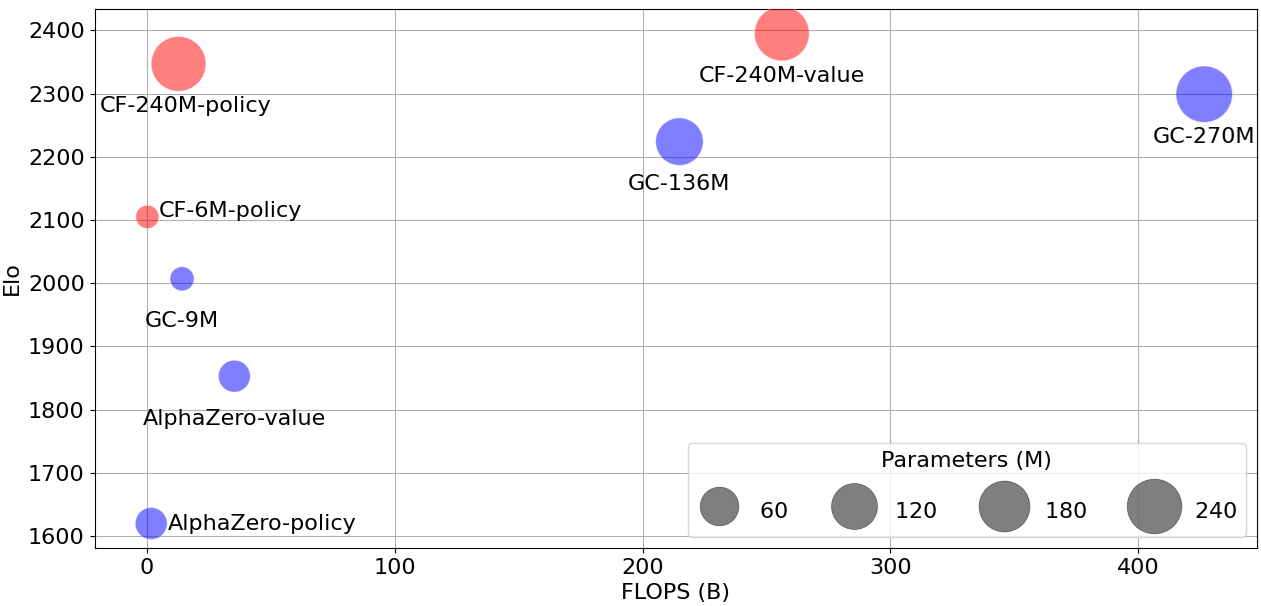}
        \end{center}
        \caption{Elo strength by floating point operations per evaluation (FLOPS) of agents constructed from our CF-6M and CF-240M models (red) against prior art (blue). Our evaluation methodology is described in \cref{sec:results}.}
                \label{fig:elo}

\end{figure*}

\section{Background}

\label{sec:background}

\subsection{Self-Attention}

\label{subsec:selfattention}

Given a sequence of tokens $(\mathbf x_1, \ldots \mathbf x_n)$ where $\mathbf x_i \in \R^{d}$, self-attention returns a sequence $(\mathbf z_1, \ldots \mathbf z_n)$ where $\mathbf z_i \in \R^{d}$. Using projection matrices $W^Q$, $W^V$, and $W^K$ in $\R^{d\times d}$, a logit $e_{ij}$ for each pair of tokens $(i,j)$ is computed through scaled dot-product attention:

\begin{equation}\label{eq:sda1}
e_{ij} = \frac{(\mathbf x_i W^Q )(\mathbf x_j W^K)^T}{\sqrt d}
\end{equation}

Attention weights are obtained via softmax:

\begin{equation}\label{eq:sda2} \alpha_{ij} = \frac{\exp(e_{ij})}{\sum_{k=1}^n \exp(e_{ik})}\end{equation}

Finally, the output is computed as a weighted sum of value information:

\begin{equation}\label{eq:sda3} \mathbf z_i = \sum_{j=1}^n \alpha_{ij} (\mathbf x_j W^V)\end{equation}

This process is repeated for each head in the self-attention layer, and the outputs are concatenated before being passed through a final linear projection.

\subsection{Position Representations}

Because self-attention is permutation-invariant, positional information must be introduced to the model through some kind of position representation. Many position representation techniques, like rotary position embeddings \cite{rope}, are designed to decay the attention weights with the Euclidean distance between tokens.

While Euclidean distance is a useful inductive bias for language and vision applications, it fails to capture the topology of the chessboard. A square with a bishop, for example, might need to attend to squares on the same diagonal rather than squares that are nearby. To demonstrate the need for more expressivity in the position representation, we compare three options: absolution position embeddings, relative biases, and the scheme of Shaw et al. \cite{shaw2018selfattention}. Changes to the vanilla self-attention formulation described in \cref{subsec:selfattention} are marked in \newfeature{blue}.

The simplest form of position representation, the absolute position embedding, adds a possibly learnable bias to each token prior to the attention layer. Calling these biases $(\mathbf c_1, \ldots \mathbf c_n)$, the absolute position embedding adds these biases to the $\mathbf x_i$ prior to the attention calculation:

\begin{equation}\label{eq:abspe}
\mathbf x_i \mapsto \mathbf x_i + \newfeature{\mathbf c_i}
\end{equation}

Unlike absolute position embeddings, relative position representations model the positions of tokens relative to each other. One simple variant introduces relative biases $d_{ij}$ which are added to the attention logits:

\begin{equation}\label{eq:biasrpe}
e_{ij} = \frac{(\mathbf x_i W^Q )(\mathbf x_j W^K)^T }{\sqrt d} \newfeature{+ d_{ij}}
\end{equation}

For one-dimensional inputs, these bias terms are shared among pairs of tokens $(i,j)$ and $(k,l)$ for which $i - j = k - l$. We adopt the two-dimensional analog of this technique, where two pairs of tokens share a relative bias if their horizontal and vertical displacements are the same.

A more general position representation introduced by Shaw et al. \cite{shaw2018selfattention}, which we adopt in our final architecture,
models the positional relationship between tokens $\mathbf x_i$ and $\mathbf x_j$ by introducing learnable vectors $a_{ij}^Q$, $a_{ij}^K$, and $a_{ij}^V$ in $\R^{d}$.  The calculation for attention logits is altered to:

\begin{equation}\label{eq:shawrpe1} e_{ij} = \frac{(\mathbf x_i W^Q \newfeature{ + a_{ij}^Q})(\mathbf x_j W^K \newfeature{ + a_{ij}^K})^T}{\sqrt d}\end{equation}

The output, meanwhile, is updated to propagate signal from $a_{ij}^V$ in addition to the output of the value projection:

\begin{equation}\label{eq:shawrpe2} \mathbf z_i = \sum_{j=1}^n \alpha_{ij} (\mathbf x_j W^V \newfeature{ + a_{ij}^V} )\end{equation}

While this technique is computationally expensive at large token counts, our models have a context length of 64, making the additional cost negligible. 




\section{Experimental Setup}

\label{sec:setup}

Our training data consists of self-play games generated with the AlphaZero process \cite{AlphaZero}. However, unlike that work, which worked in the reinforcement learning setting, we create a static dataset of self-play games from an older reinforcement learning run. That run used a transformer model of roughly 100 million parameters at 600 nodes per move which was almost identical in architecture to our main runs, barring its use of a slightly weaker position representation than that of our final architecture. We therefore work in the supervised setting without the need for online data generation, allowing for much faster training. Initial experiments showed no quality degradation on a static dataset.

All models we train employ a standard encoder-only backbone with a context of 64 tokens corresponding to the squares on the chessboard, with one of the position representations described in \cref{sec:background}. The training targets consist of the main policy and value targets in addition to a mix of auxiliary targets, and we use the Nadam optimzer with $\beta_1 = 0.9$, $\beta=0.98$, $\epsilon=10^{-7}$, and gradient clipping 10. The final model checkpoints used for evaluation are generated with stochastic weight averaging \cite{swa}. A more detailed description of the architecture and losses is provided in the Appendix.


Our largest model, CF-240M, has 15 encoder layers with an embedding depth of 1024, a head count of 32, and a feedforward depth of 4096, for a total of 243 million parameters. It was trained for 3.7 million steps with a batch size of 4096 on 8 A100 GPUs using data parallelism. The learning rate was initialized at $1 * 10^{-3}$ and manually reduced to $3 * 10^{-4}$ at 3.2 million steps and $1 * 10^{-4}$ at 3.6 million steps. The dataset consisted of 500 million games generated from mid-2023 to mid-2024, with each game containing roughly 200 positions.

We also train a smaller model, CF-6M, with 8 encoder layers, an embedding depth of 256, a head count of 8, and a feedforward depth of 256, for a total of 6 million parameters. The same configuration is used for ablations of the position representation. CF-6M and its ablations were each trained on a single A100 GPU with a batch size of 2048. The learning rate was initialized at $5 * 10^{-4}$ and reduced to $1.58 * 10^{-4}$ and $5 * 10^{-5}$ at 1.6 and 1.8 million steps, respectively. The dataset consisted of 53 million games generated in April 2024. None of our training runs exhibited overfitting, which is likely due to the size of the datasets relative to the parameter counts of the models.

\section{Results}

\label{sec:results}

We compare the playing strength and puzzle-solving ability of agents constructed from our models to prior work and ablate the position representation in the attention mechanism to demonstrate the criticality of accurately modeling positional information.

We include in our analysis two types of agents: those that function based on policy information by picking the move which is ranked highest in the policy vector predicted by the model, and those that function based on value information, emulating a search of depth 1 by evaluating the model for each legal move and selecting the move that maximizes the position evaluation. The policy strategy requires a single model evaluation, while the value maximization strategy requires an evaluation for each legal move. To estimate the floating point operations per evaluation (FLOPS) used by an agent of the value maximization type, we multiply the model FLOPS by 20, which is a rough estimate of the average number of legal moves available in a position.

We construct agents from our models with both strategies, denoting an agent by its model name followed by the strategy it uses (e.g., CF-240M-policy and CF-240M-value). Policy and value agents are constructed from the AlphaZero model in the same way. Our analysis also includes models from Ruoss et al. \cite{ruoss2024grandmasterlevel} which use the value maximization approach. We use the final checkpoints of their main runs having parameter counts of 9 million, 136 million, and 270 million, referring to these as GC-9M, GC-136M, and GC-270M, respectively.


\subsection{Playing Strength}

To estimate the playing strength of our agents, we play 1,000 games with them against the GC-270M agent of Ruoss et al. \cite{ruoss2024grandmasterlevel}, tying the Elo rating of GC-270M to the value reported in that paper. We use the same opening book and testing configuration as that work to achieve a similar comparison to theirs. Elo values for the GC and AlphaZero agents are taken from that paper. The results are shown in \cref{table:main} and \cref{fig:elo}.

\subsection{Puzzles}

\begin{figure*}[!h]
        \begin{center}
        \includegraphics[width=\textwidth]{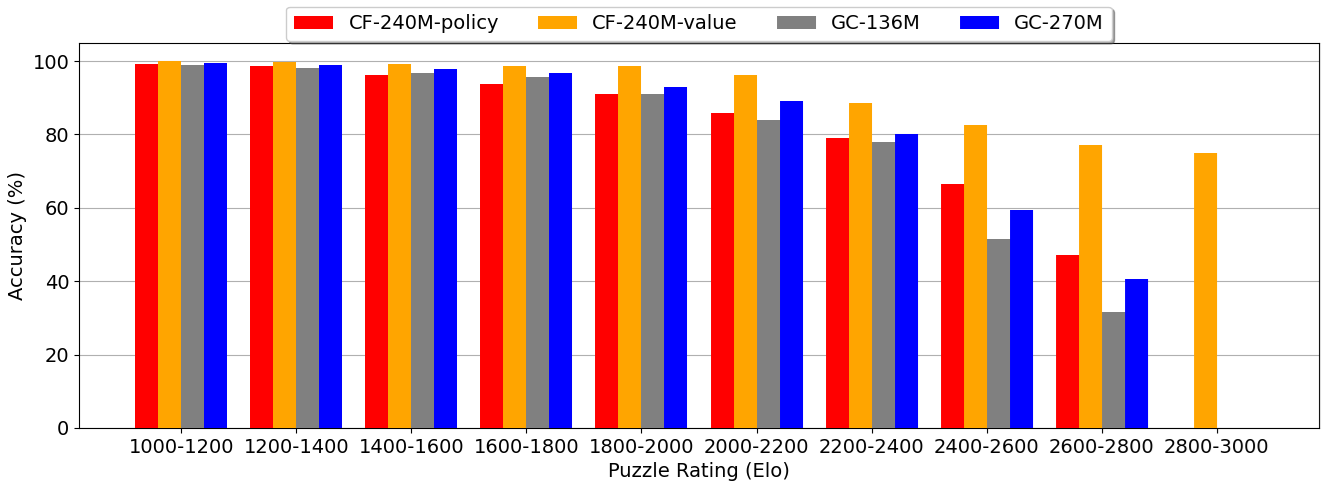}
        \end{center}
        \caption{Puzzle-solving ability on puzzles rated 1000-3000 of our CF-240M-policy and CF-240M-value agents against the GC-136M and GC-270M agents of Ruoss et al. \cite{ruoss2024grandmasterlevel}.}
                \label{fig:puzzles}

\end{figure*}

We also evaluate the puzzle-solving ability of our agents on the puzzle database curated by Ruoss et al. \cite{ruoss2024grandmasterlevel}. Our evaluation follows the same methodology as that paper, with an agent deemed to solve a puzzle only if it correctly picks each move in the sequence of the solution. The results are shown in \cref{table:main}. \cref{fig:puzzles} graphs the accuracy of our two CF-240M agents on this test set against the GC-136M and GC-270M agents of Ruoss et al. \cite{ruoss2024grandmasterlevel}.

\subsection{Ablations}

We ablate the position representation of the CF-6M model, comparing the final test metrics of the three position representations described in \cref{sec:background} on a held-out test set of around 1 million games sharded from the training dataset. The reported statistics are for the main policy and value targets. As shown in \cref{table:ablate}, our chosen technique, that of Shaw et al. \cite{shaw2018selfattention}, substantially outperforms both the relative bias and absolute position embedding representations. At this parameter count, a doubling in model size yields an increase in policy accuracy of around 1.5\%. The Shaw encoding outperforms the learned absolute position embedding by 1.83\% policy accuracy, suggesting that a good choice of position representation can in large part replace the need for model scale.


\begin{table}[h!]

\centering

\caption{Results for Ablating Position Representation}
\label{table:ablate}

\begin{tabular}{@{}lrrrr@{}}
\toprule

&  \multicolumn{2}{|c|}{\textbf{Loss}}   & \multicolumn{2}{|c|}{\textbf{Accuracy (\%)}} \\
\midrule      
\textbf{Representation} & \textbf{Policy} & \textbf{Value} & \textbf{Policy} & \textbf{Value}  \\
\midrule                                                                                                
Absolute & 0.3460 & 0.5607 & 57.44 & 89.11  \\

\midrule

Relative bias & 0.3321 & 0.5586 & 58.23 & 89.26  \\

\midrule                                                                                       
Shaw et al. \cite{shaw2018selfattention} & \textbf{0.313} & \textbf{0.5549} & \textbf{59.27} & \textbf{89.53} \\

\midrule  

\end{tabular}

\end{table}

\subsection{Analysis}

The main results can be found in \cref{table:main}. Our agents consistently outperform prior work in both puzzle-solving ability and Elo performance at a fraction of the computational cost. Our CF-6M-policy agent outperforms the AlphaZero-policy agent in both metrics at 8x fewer FLOPS. At 30x fewer FLOPS, our CF-240M-policy agent matches the puzzle performance and exceeds the Elo performance of the grandmaster-level agent GC-270M of Ruoss et al. \cite{ruoss2024grandmasterlevel}.

\begin{table}[!ht]

\centering
\caption{Comparison of Playing Strength and Puzzle Solving Ability}
\label{table:main}
\begin{tabular}{@{}lrrrrrrrrr@{}}

\toprule

\textbf{Agent} & \textbf{Elo} & \textbf{Puzzles (\%)}  & \textbf{FLOPS} \\

\midrule
CF-6M-policy (ours)  &2105  (±28)& 65.3 & 214M \\
CF-240M-policy (ours)    &2347  (±10) & 93.5 & 12.8B    \\
CF-240M-value (ours) &2385  (±10)  &  97.6 & 256B     \\

\midrule     
GC-9M  & 2007 (±15) & 85.5 & 14.2B       \\
GC-136M  & 2224 (±14) & 92.1 & 215B \\
GC-270M &  2299 (±14)  & 93.5 &   427B  \\

\midrule                     
AlphaZero-policy   & 1620 (±22) & 61.0 & 1.77B   \\
AlphaZero-value   & 1853 (±16) &  82.1 & 35.3B  \\

\midrule

\end{tabular}
\end{table}


\section{Attention Maps}

\label{sec:attmaps}

\begin{figure}[!h]
    \centering
    \includegraphics[width=0.49\columnwidth]{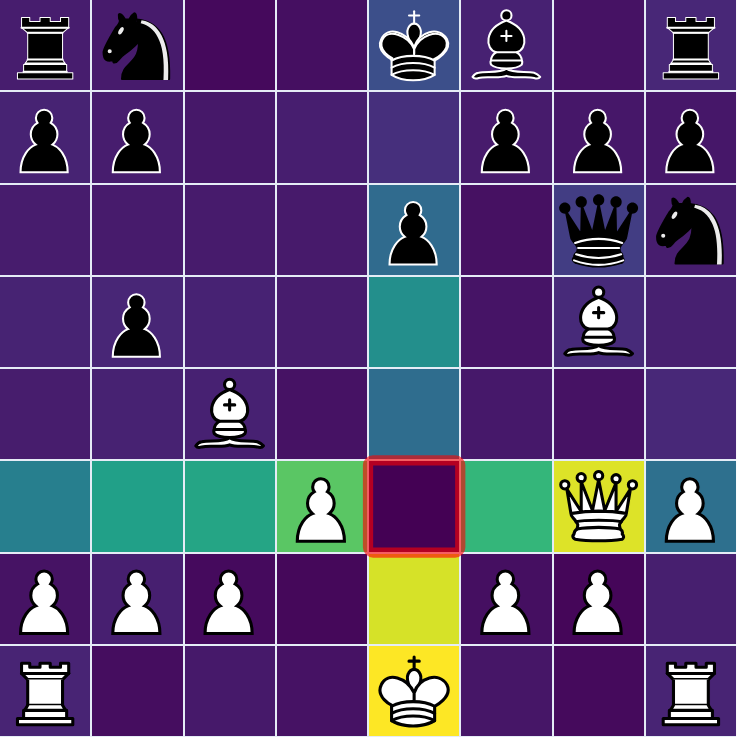}
    \includegraphics[width=0.49\columnwidth]{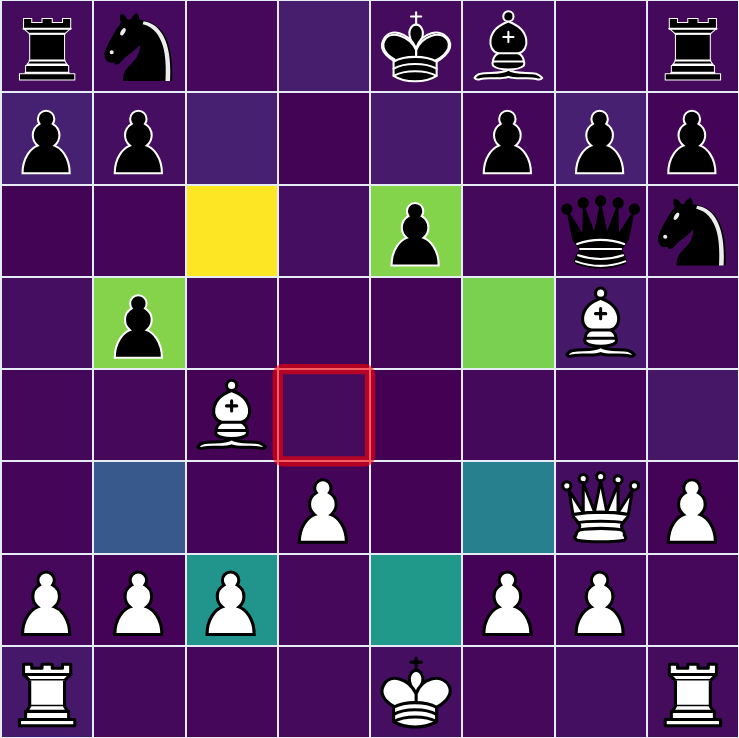}
    
    \vspace{5px}
    \includegraphics[width=0.49\columnwidth]{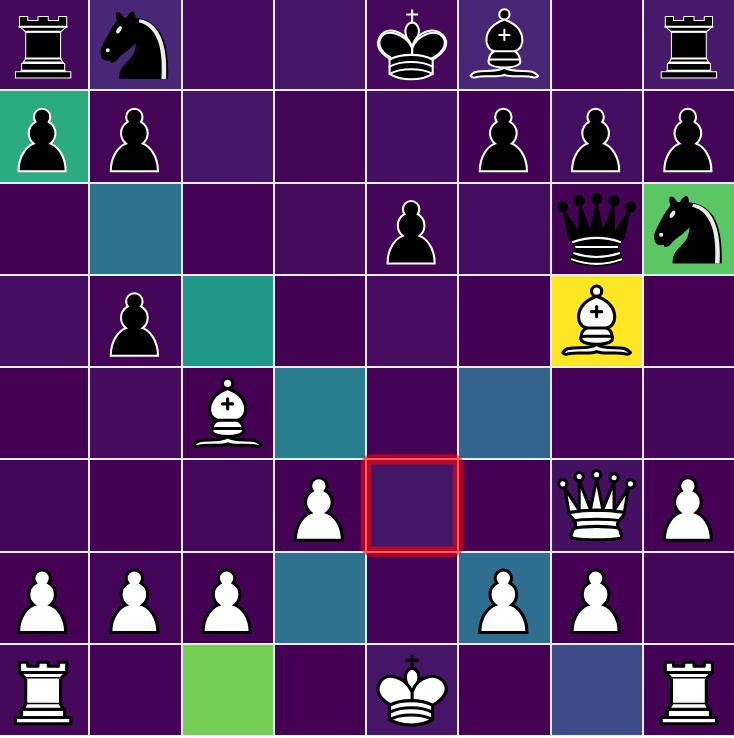}
    \includegraphics[width=0.49\columnwidth]{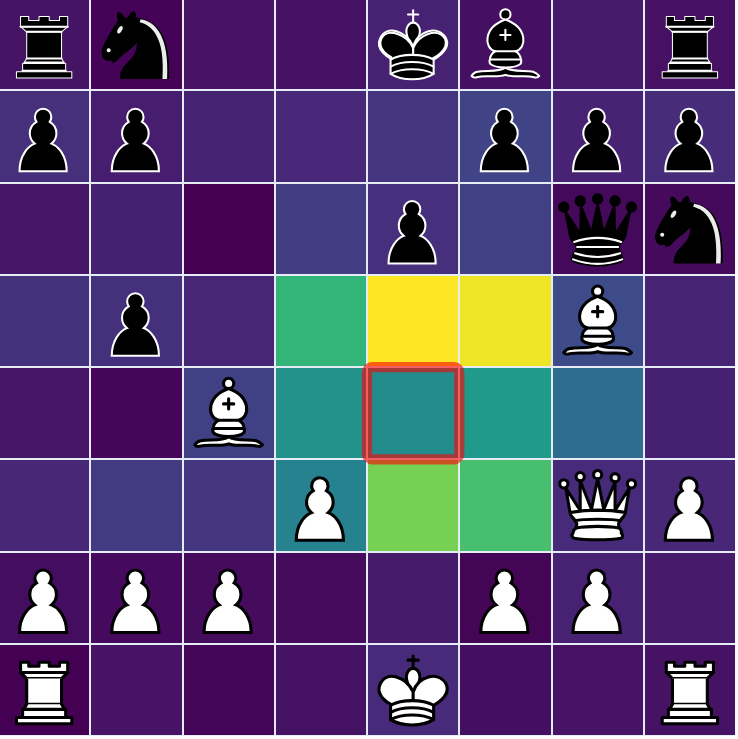}

    \caption{Attention maps of heads corresponding to the movement of a particular piece. The square highlighted in red is the one producing the query. }
        \label{fig:pmheads}

\end{figure}

\begin{figure}[!h]

    \centering
    \includegraphics[width=0.49\columnwidth]{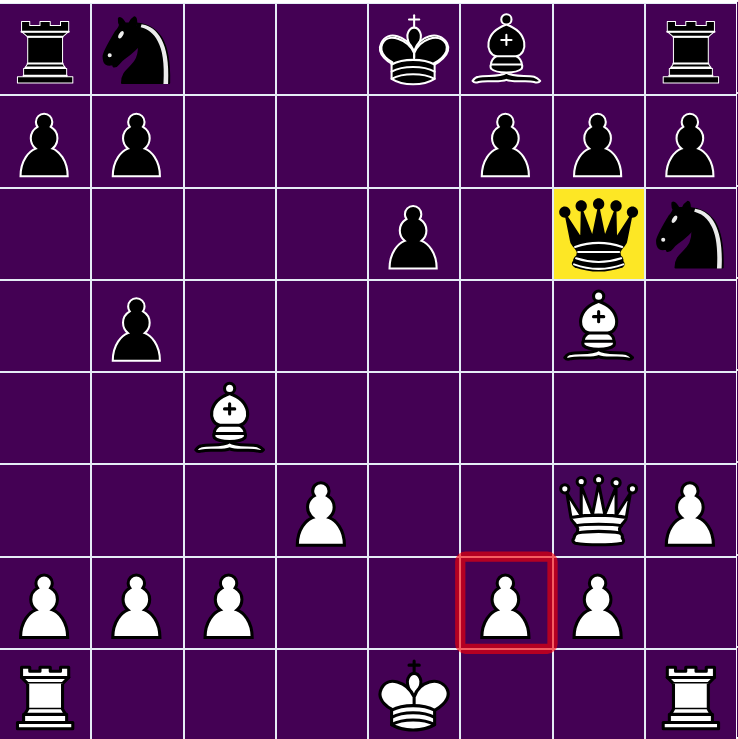}
    \includegraphics[width=0.49\columnwidth]{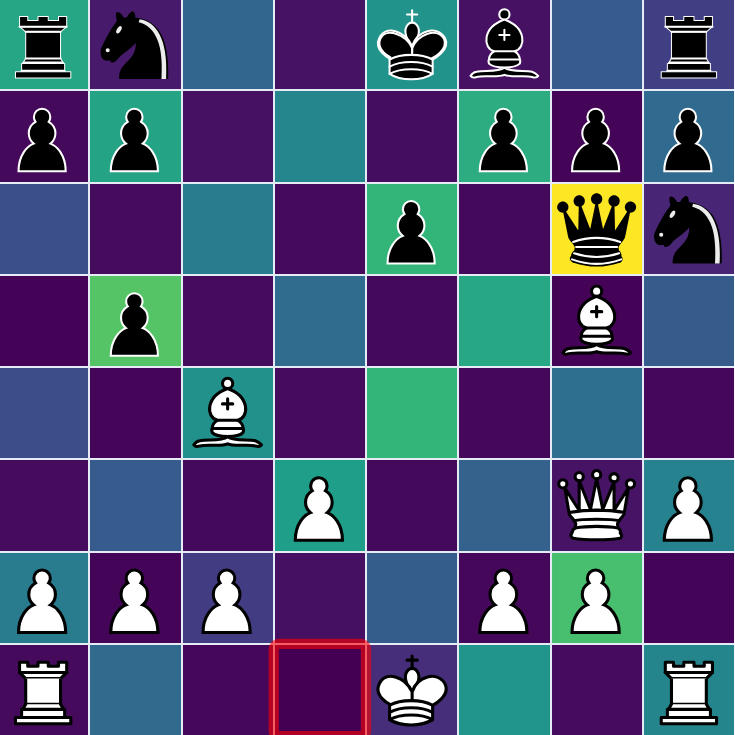}
    
    \vspace{5px}
    \includegraphics[width=0.49\columnwidth]{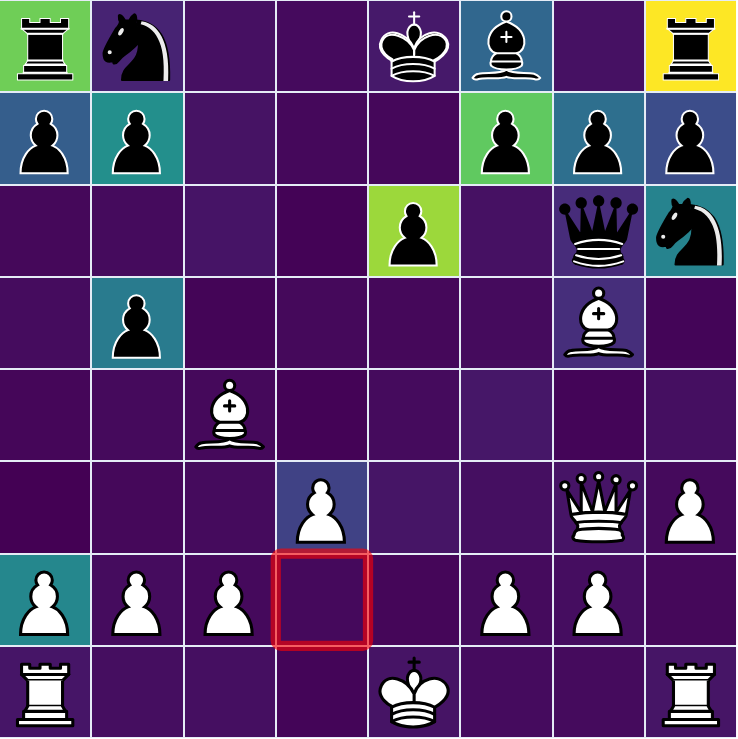}
    \includegraphics[width=0.49\columnwidth]{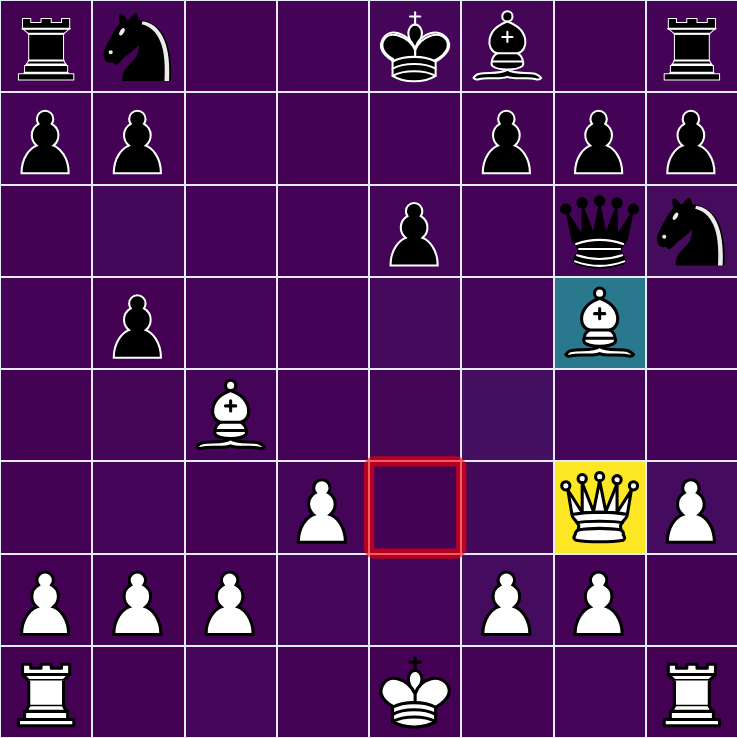}

    \caption{Attention maps of several additional heads with easily interpretable patterns. The square highlighted in red is the one producing the query.}
        \label{fig:specheads}

\end{figure}

One advantage of transformers is the interpretability of their attention maps. We generate visualizations\footnote{Code for generating attention visualizations is available at \blackurl{https://github.com/Ergodice/lc0-attention-visualizer}.} of the attention maps of our CF-240M model, which we plot in \cref{fig:pmheads} and \cref{fig:specheads}. We select a querying square and color squares based on their attention weights for the querying square. Small weights are colored purple and large weights are colored yellow.

The majority of attention heads in our models represent the movement of a particular piece, attending to squares which are a bishop's, knight's, rook's, or king's move away, depending on the specialization of the head. We give four examples from CF-240M in \cref{fig:pmheads}. From left to right then top to bottom, the heads shown are layer 5 head 24, layer 1 head 28, layer 1 head 14, and layer 15 head 22. The attention maps are fairly static across positions, always retaining their specialization even if the piece whose movement type they specialize in is no longer on the board.

Interspersed with these piece movement heads are several other heads with recognizable patterns. We give four examples in \cref{fig:specheads}, describing them from left to right then top to bottom. Head 3 in layer 1 always attends to the opponent's queen. Head 29 in layer 2 attends to squares which are of the same color on the checkerboard as the querying square. Head 14 in layer 15 attends to squares occupied by the opponent's pieces with roughly equal weights. Finally, head 2 in layer 11 attends to pieces that can move to the querying square. This final type of head appears across model scales, though always in the second half of the encoder layers, suggesting that they rely on information about piece movement accumulated in early layers.

The variety of these miscellaneous heads suggests that the model develops a broad view of chess principles. The same-checkerboard-color head, for example, may provide information about the maneuverability of bishops. The head attending to the opponent's pieces may be estimating the material difference between the players or gleaning the game stage, i.e., opening, middlegame, or endgame, from the types of pieces on the board.

Unlike in vision and language applications, proximity of tokens seems to have little effect on attention weights, confirming the importance of a position representation which can adapt to the topology of the chessboard. Another notable feature of the attention maps is that tokens rarely attend to themselves. This is in contrast to language models, where it is quite common \cite{vaswani2017attention}.

\section{Playstyle}

\label{sec:playstyle}

\begin{figure}[!h]

    \begin{center}
        \begin{center}
        \includegraphics[width=0.5\columnwidth]{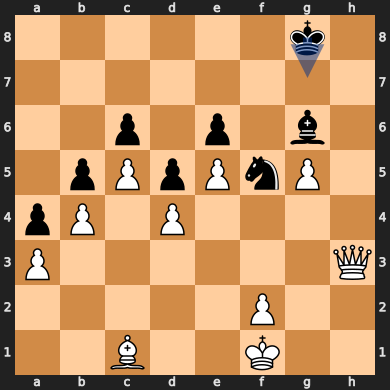}
            \end{center}

        Position 1: fortress detection. CF-240M finds the only drawing move, Kg7, and evaluates the position as drawn.
        \hfill
        
            \begin{center}
            \includegraphics[width=0.49\columnwidth]{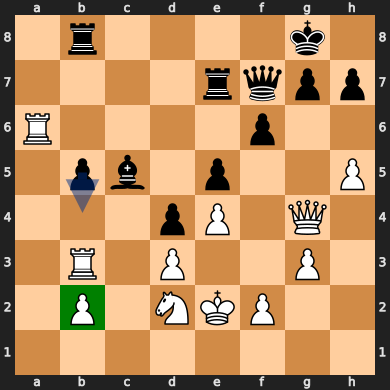}
            \includegraphics[width=0.49\columnwidth]{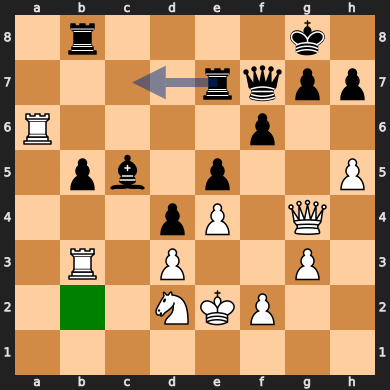}
            \end{center}

        \vspace*{-2mm}

        Position 2: trapped piece detection. If white has a pawn on b2, CF-240M opts to trap the white rook on b3 by pushing b4. However, if the pawn on b2 is not present, it abandons this idea.

        \begin{center}
        \includegraphics[width=0.5\columnwidth]{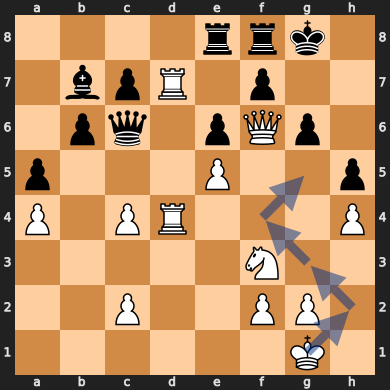}
        \end{center}

        \vspace*{-2mm}

        Position 3: long-term planning. CF-240M finds a famous king walk that was played out by Nigel Short.

    \end{center}

    \caption{Positions in which our models exhibit a humanlike understanding of the game, detecting positional ideas that elude top minimax-based engines.}
        
    \label{fig:pos}

\end{figure}

Our largest model, CF-240M, appears to have an understanding of chess which is more akin to humans than that of conventional engines. In particular, it is capable of detecting difficult positional motifs that top chess engines struggle with. One position demonstrating this humanlike game understanding arose between Lc0 and Stockfish in Game 189 of the 23rd Computer Chess Championship Rapid Tournament,\footnote{See \blackurl{https://www.chess.com/computer-chess-championship\#event=ccc23-rapid-finals\&game=189}.} where the former used a transformer with 190 million parameters trained by the authors. Lc0 managed to construct a fortress from a lost position, leading the game into a draw. A position from this game is shown as Position 1 in \cref{fig:pos}. CF-240M assigns the only drawing move, Kg7, a policy of 92.8\%, and evaluates the position as drawn \wdl{2.2}{76.3}{21.5}. Stockfish assigned the same position a +7 evaluation after searching 11 billion nodes, and its evaluation stayed above +3 for over 100 moves. Stockfish evaluations not directly interpretable as win probabilities, but a +3 evaluation corresponds roughly to an over 99\% probability of winning.

Another position demonstrating this humanlike positional understanding arose between Lc0 and Stockfish in Game 163 of the 21st Computer Chess Championship Rapid Tournament,\footnote{See \blackurl{https://www.chess.com/computer-chess-championship\#event=ccc21-rapid-semifinals\&game=163}.} where Lc0 used a transformer of around 100 million parameters trained by the authors. Shown as Position 2 in \cref{fig:pos} is a critical moment in the game when Lc0 played b4, trapping the white rook on b3. CF-240M assigns this move the highest policy at 37\% and evaluates the position as slightly winning \wdl{40.3}{54.1}{5.6}. Lc0 ended up winning the game, with Stockfish's evaluation only going negative several moves later despite searching hundreds of millions of positions.

If the pawn on b2 is removed and this trapping idea is no longer feasible, the evaluation becomes much more drawish \wdl{17.1}{77.7}{5.2}, despite black gaining a material advantage. Additionally, the model loses interest in b4, with the policy for that move decreasing to 4.76\% and Rc7 becoming the favored move.

The positional understanding of our CF-240M model seems to extend to long-term planning as well. As shown in Position 3 in \cref{fig:pos}, CF-240M picks the right move at each of white's turns in a famous king walk played by GM Nigel Short as white against GM Jan Timman. The line continues 31. Kh2 Rc8 32. Kg3 Rce8 33. Kf4 Bc8 34. Kg5.

\section{Related Work}

\label{sec:related}

Independent work in the same vein as this paper by Ruoss et al. \cite{ruoss2024grandmasterlevel} trains transformers at scale for chess, demonstrating that a transformer-based approximation of a strong oracle can achieve grandmaster-level strength without search. Unlike this paper, they focus on distilling a search-based algorithm into a transformer model rather than optimizing for playing strength.

Czech et al. \cite{czech2023representation} design a lightweight transformer block for chess and show that the performance of neural chess models can be improved with carefully chosen input representations and value targets. Similar work in Go \cite{wu2020accelerating}, \cite{wu2024} adds auxiliary training targets and enhanced input representations to accelerate learning in the AlphaZero process.

Other work interprets the inner workings of neural chess models through ablations, probing of internal states, and analysis of attention maps. McGrath et al. \cite{mcgrath2021acquisition} analyze the acquisition of chess knowledge by AlphaZero during training, while Jenner et al. \cite{lookahead} examine the role of attention maps and lookahead in an older iteration of transformer models trained by the authors.
\section{Conclusion}

\label{sec:conclusion}

Though the vanilla transformer is an effective generalist architecture, as we have shown, there can be substantial benefit to exploring domain-specific enhancements and inductive biases like effective position representations. This holds true in resource-limited scenarios but especially in search-based applications like computer chess where speed and accuracy are interchangeable. This idea is exemplified by work such as AlphaFold \cite{alphafold}, which predicts the structure of proteins by modeling them with a graph where edges correspond to relationships between residues.

The fungibility of quantity and quality has featured prominently in recent work aiming to improve the reasoning capabilities of language models by encouraging them to delineate their thought process before giving a final answer. For example, chain-of-thought prompting \cite{cot} prompts a language model with example reasoning paths, increasing accuracy at the expense of additional computational cost. OpenAI's o1 model \cite{OpenAIo1} takes this idea further and is trained to perform a large reasoning step before giving its final answer. In the domain of automated theorem-proving, AlphaProof \cite{AlphaProof} solves olympiad-level math problems by searching through potential proofs with a model specialized for theorem proving.

We hope our work, which demonstrates the potential usefulness of domain-specific enhancements and shows that in some cases deep learning can succeed where search fails, can provide lessons for this new search-based reasoning paradigm.

\begin{appendices}

\label{sec:appendix}

\section{Model Architecture}

\label{sec:architecture}

\subsection{Input Encoding}

The input to our network is a sequence of 64 tokens: one token for each square on the board read from left to right then bottom to top. The board is flipped with the side to move. Each input token has length 112 and consists of a concatenation of:

\begin{itemize}
\item 8 one-hot vectors of length 12 describing the piece at that square for the current position and past 7 positions.
\item En passant and castling information.
\item The number of positions since the last capture, pawn move, or castle, divided by 100.
\item Whether each of the current and past 7 positions is a repetition.
\end{itemize}

To generate token embeddings, we apply a linear projection to the input tokens, then add and multiply by learned offset vectors which are separate across tokens and depth. This gives the model absolute positional information and was found in initial experiments to incrementally improve model quality.

\subsection{Body}

The body of our models consists of a stack of encoder layers with Post-LN normalization and the initialization/gain scheme DeepNorm \cite{wang2022deepnet}. We use a fixed context length of 64. We use Mish \cite{misra2019mish} activations in the feedforward sublayer, and all projections in the output heads are followed by this activation unless otherwise stated.

Following \cite{palm}, we omit biases in the QKV projections and omit centering and biases in the encoder normalization layers, finding this to increase training throughput by around 10\% without degrading model quality. However, we retain biases in the feedforward and post-attention projections since removing them degraded quality in initial experiments.

\subsection{Output Heads}

Similar to AlphaZero, our models have heads for policy and value predictions. However, we add several auxiliary policy and value targets to increase convergence speed, following research in Go \cite{wu2020accelerating}, \cite{wu2024}. 

\textbf{Policy Heads} We introduce a new policy head based on a modified self-attention mechanism. Moves are encoded by the starting square and destination square of the piece moved. Given the sequence of tokens outputted by the body, we generate policy embeddings by applying a dense layer of depth equal to the model's embedding size. From this we generate, via linear projection, a set of query vectors corresponding to the starting square and a set of key vectors corresponding to the destination square, both with depth equal to the depth of the encoder body. 

Logits for moves are calculated via scaled dot product as described in \cref{eq:sda1}. The result is a 64x64 matrix representing all possible traversals from one square on the chessboard to another. These traversals are sufficient to represent all moves that can occur within the rules of chess, with the exception of promotions. When a pawn advances to the last rank of the board, it must be promoted to a knight, bishop, rook, or queen. To represent these special moves, we apply a linear projection to the key vectors representing the promotion rank, generating an additive bias for each possible promotion piece. This bias is then applied to the logits representing all possible traversals between the penultimate rank and the promotion rank to generate additional logits for each possible promotion. 

To generate the final policy vector, we apply a softmax over all logits, masking illegal moves to increase training stability. We did not see performance gain from using any of the position representations described in \cref{sec:background}, likely because the embedding size is much larger than the token count.  

\textbf{Value Heads}
To model value information, we apply a linear projection of depth $d_\text{value}$ to the output of the body, where $d_\text{value} = 32$ by default. We then flatten the result and apply another projection of size 128, which we call the value embedding. Each of our models has three value heads, each of which generates its own value embedding separately from the body output. Each of these value heads uses this embedding to predict one or more training targets.

The ``result'' value head predicts the result of the game from among win, draw, and loss with cross-entropy loss. The ``q'' and ``short-term'' value heads each predict three targets: a reward trained with L2 loss, a categorical distribution over rewards trained with cross-entropy loss, and the error between the predicted and true reward. Respectively, these latter two heads use the reward estimate produced during self-play and an exponential moving average of future rewards with expected depth 6. The gradient of the reward prediction is detached when calculating the error loss to prevent the error head from affecting the reward prediction.



\section{Train Losses}
\label{sec:losses}

As noted above, our models train on several auxiliary targets, which were found to accelerate model convergence and improve final performance in Go \cite{wu2020accelerating}, \cite{wu2024}. Here we describe the losses for the main and auxiliary targets. The final loss function is a weighted sum of these terms.

\textbf{Policy Targets} 

\begin{itemize}
  \item Vanilla policy head: predicts the policy target produced by self-play.

  \[-c_{\text{pol}} \sum_{m \in \text{moves}} \pi(m)\log(\hat \pi(m)) \]

  where $\pi$ is the policy target, $\hat \pi$ is the model's prediction of $\pi$, and $c_{\text{pol}} = 1$.
  
  \item Soft policy head: predicts a high-temperature version of the policy target.
  
  \[-c_{\text{softpol}} \sum_{m \in \text{moves}} \pi_{\text{soft}}(m)\log(\hat \pi_{\text{soft}}(m)), \]

  where $\pi_{\text{soft}}$ is taken from $\pi$ by setting the temperature to 4 and $\hat \pi_{\text{soft}}$ is the model's prediction of $\pi_{\text{soft}}$. We set $c_{\text{softpol}} = 8$, though this term is still small compared to the main policy loss because the loss is muted by the high temperature.
\end{itemize}

\textbf{Value Targets} 
    \begin{itemize}
      \item Game result: predicts the outcome of the game.

      \[c_{\text{value-wdl}} \sum_{r \in \{\text{win, loss, draw}\} } v(r)\log(\hat v(r)) \]

      where $v$ is the game result, $\hat v$ is the model's prediction of $v$, and $c_{\text{value-wdl}} = 1$.

      \item L2 value: predicts a scalar reward produced by self-play.

        \[c_{\text{value-l2}} (q - \hat q)^2\]

        where $q$ is the true reward predicted by $\hat q$ and $c_{\text{value-l2}}=1$.

      \item Categorical value: predicts a categorical distribution over reward estimates.
      \[c_{\text{value-cat}} \sum_{x \in \text{score buckets} } z(x)\log(\hat z(x))\]

      where $z$ is a one-hot vector representing the bucket into which the recorded reward $q$ falls into. We set $c_{\text{value-cat}} = 0.1$ for CF-240M and excluded this target for all other training runs. 

      \item Value error: predicts the error of the L2 value prediction.
            \[c_{\text{value-error}} (\hat e - (q - \hat q)^2)^2  \]

        where $c_{\text{value-error}}$ = 1 and $\hat e$ is the model's prediction of the squared difference between the predicted and true reward.

   \end{itemize}

\section{Other Techniques Tried}
\label{sec:other}

In addition to the techniques reported here, we tried several architectural modifications that did not bear fruit. For example, replacing the feed-forward layer with a Gated Linear Unit \cite{glu} slightly degraded quality at constant model FLOPS. Using a sparsely gated \cite{moe} or soft \cite{softmoe} mixture of experts in place of the feed-forward layer also did not improve performance.

\end{appendices}

\section*{Acknowledgment}

\label{sec:acknowledgments}

This paper relied on extensive contributions from the Lc0 community. We are grateful to Twitch user KittenKaboodle for generously providing the hardware on which experiments were run, and to Github user jkormu for implementing the attention visualization. We are also thankful to David Elliott, Hunter Monroe, Denys Ivanov, and Evan Engler for proofreading and providing insight. Any errors that remain are our own.

Training code and templates for \cref{fig:elo}, \cref{fig:puzzles}, and \cref{fig:pos} were generated with help from Github Copilot, and much of the text was checked for spelling and writing quality with GPT-4.

\bibliographystyle{IEEEtran}
\bibliography{references}



\end{document}